\documentclass[10pt,twocolumn,letterpaper]{article}

\usepackage[pagenumbers]{cvpr} 

\usepackage{multirow} 
\usepackage[dvipsnames]{xcolor}

\usepackage[symbol]{footmisc}

\definecolor{cvprblue}{rgb}{0.21,0.49,0.74}
\usepackage[pagebackref,breaklinks,colorlinks,citecolor=cvprblue]{hyperref}

\def\paperID{13948} 
\def\confName{CVPR}
\def\confYear{2024}

\title{Dual-Modal Prompting for Sketch-Based Image Retrieval}

\author{Liying Gao$^{1\ast}$ 
\quad 
Bingliang Jiao$^{1\ast}$ 
\quad Peng Wang$^{1\dagger}$ \\ \quad Shizhou Zhang$^{1}$ \quad Hanwang Zhang$^2$ \quad Yanning Zhang$^1$ \vspace{0.3em} \\
{\normalsize $^1$Northwestern Polytechnical University } \quad
{\normalsize $^2$ Nanyang Technological University} \quad \\
{\tt\small {gaoliying,~bingliang.jiao}@mail.nwpu.edu.cn, peng.wang@nwpu.edu.cn}
}

\begin{document}
\maketitle

\footnotetext[1]{Equal contributions. }
\footnotetext[2]{Corresponding Author.}

\begin{abstract}
Sketch-based image retrieval (SBIR) associates hand-drawn sketches with their corresponding realistic images. In this study, we aim to tackle two major challenges of this task simultaneously: i) \textbf{zero-shot}, dealing with unseen categories, and ii) \textbf{fine-grained}, referring to intra-category instance-level retrieval. Our key innovation lies in the realization that solely addressing this cross-category and fine-grained recognition task from the generalization perspective may be inadequate since the knowledge accumulated from limited seen categories might not be fully valuable or transferable to unseen target categories. Inspired by this, in this work, we propose a dual-modal prompting CLIP (DP-CLIP) network, in which an adaptive prompting strategy is designed. Specifically, to facilitate the adaptation of our DP-CLIP toward unpredictable target categories, we employ a set of images within the target category and the textual category label to respectively construct a set of category-adaptive prompt tokens and channel scales. By integrating the generated guidance, DP-CLIP could gain valuable category-centric insights, efficiently adapting to novel categories and capturing unique discriminative clues for effective retrieval within each target category. With these designs, our DP-CLIP outperforms the state-of-the-art fine-grained zero-shot SBIR method by $7.3\%$ in Acc.@1 on the Sketchy dataset. Meanwhile, in the other two category-level zero-shot SBIR benchmarks, our method also achieves promising performance.


\end{abstract}    
\section{Introduction}
Sketch-based image retrieval (SBIR)~\cite{Collomosse_2019_CVPR,xu2018sketchmate,guo2017sketch} is a crucial problem of sketch understanding, which aims to match corresponding realistic images with given hand-drawn sketches. Given the scarcity and limited variety of sketches, in real-world applications, the SBIR task inevitably encounters unseen sketch categories.  Consequently, the research efforts have majorly centered on addressing the zero-shot sketch-based image retrieval (ZS-SBIR) problem~\cite{yelamarthi2018zero,dey2019doodle}.

\begin{figure}[]
	\begin{center}\includegraphics[width=1.0\linewidth]{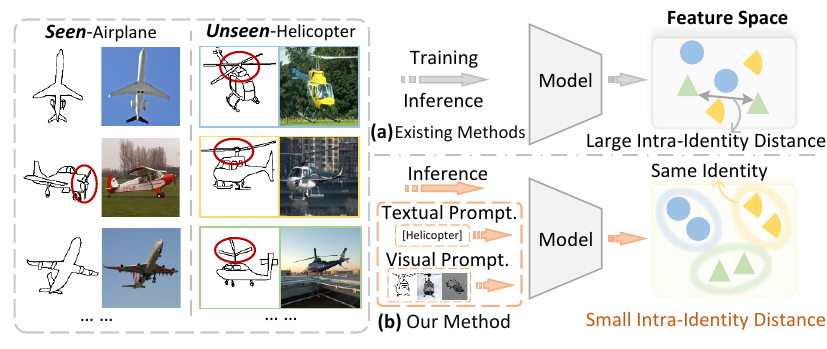}
	\end{center}
	\vspace{-0.7cm}
	\caption{ 
        The comparison between (a) existing methods and (b) our method in addressing the fine-grained zero-shot SBIR task. Most existing models accumulate knowledge from limited seen categories and directly transfer it to unseen ones.
        We believe this is sub-optimal since not all knowledge acquired in seen categories is transferable, some of which could be invalid or even detrimental.
        For instance, the presence of the propeller (marked by red bounding boxes) may serve as a distinctive feature for identifying airplanes but not for helicopters.
        To solve this, we introduce a dual-modal prompting strategy. In this strategy, we employ the corresponding textual category label and a set of images from the target category to equip the model with category-centric insights.
        This could prompt it to adapt to the target category, thereby achieving better retrieval.
		}
    \vspace{-0.5cm}
	\label{fig:intro}
\end{figure}

Most existing ZS-SBIR works have concentrated on category-level retrieval, aiming to match sketches and photos that belong to the same category. While significant progress has been achieved~\cite{sain2022sketch3t,tian2022tvt, wang2022prototype,lin2023explainable}, this coarse-grained recognition might not fully satisfy the practical needs of real-world ZS-SBIR applications. For an intuitive example, when searching for lost items, we hope to match the exact corresponding instance with our hand-drawn sketch. Recently, the fine-grained ZS-SBIR task has been proposed~\cite{pang2019FG_CC,sain2023clip}, which aims to address this gap by achieving instance-level matching between sketches and realistic images. This fine-grained matching introduces new challenges to the ZS-SBIR task and increases its applicability, encouraging us to delve into this field of research.

In this paper, our key insight is that the generalization learning approaches in previous ZS-SBIR research, which derive knowledge that is effective for handling seen categories from training set~\cite{pang2019FG_CC} or pre-trained models~\cite{sain2023clip} and then
directly transfer it to target unseen categories, are not apt for this fine-grained and zero-shot recognition scenario. This is primarily because the knowledge acquired for distinguishing individuals within seen categories may not be entirely transferable or applicable to unseen categories. 
For instance, as illustrated in Figure~\ref{fig:intro}, the presence of the propeller may serve as a distinctive feature for identifying airplanes but not for helicopters. This necessitates the retrieval model to possess ample adaptability, enabling it to selectively utilize and adjust its accumulated knowledge to adapt to the target categories.

To achieve this goal, we introduce an adaptive prompting strategy tailored for the fine-grained ZS-SBIR task. In our approach, we provide the retrieval model with the textual category label and a collection of images within the target category as support data. Leveraging this support data, we train the model to extract category-centric knowledge,  enabling efficient adaptation to target categories. 
Practically, in this work, we propose a dual-modal prompting CLIP (DP-CLIP) network for fine-grained ZS-SBIR. 
We employ CLIP~\cite{radford2021clip} as the backbone model, leveraging its extensive knowledge and remarkable zero-shot capabilities acquired from diverse pre-training data. Furthermore, in our DP-CLIP framework, we design a visual prompting module and a textual prompting module. The objective of the visual prompting module is to learn category-specific insights from a set of support images within the target category, facilitating the adaptation of DP-CLIP to the target category. In this step, we notice the visual prompt tuning~\cite{jia2022vptl} could be a good tool, which guides the deep model to adapt to target tasks via injecting it with task-specific prompt tokens. Based on this idea, in our visual prompting module, we use a transformer layer to extract and transfer category knowledge from given support images into a few learnable tokens. These tokens are then employed as category-specific prompts to adapt our model to the target category. 
Notably, our visual prompting module only requires a few (three) support images per category, and it also works well when learning category insights from the input image itself.

For the textual prompting module, our objective is to derive knowledge about the target category from the CLIP's textual encoder and leverage it to guide the adaptation. To achieve this, we employ the textual encoder of CLIP to embed the target category label into textual features. Subsequently, we utilize these textual features to produce a series of category-specific channel scaling vectors. These scaling vectors are then applied to the hidden features of the CLIP visual encoder, directing the model to focus on channels that are relevant to the current category. In addition to the dual-modal prompting modules, we employ a customized patch-level matching module in our DP-CLIP to capture detailed correspondences between sketches and photos.

The main contributions of this paper can be summarised as follows:
1) We propose a dual-modal prompting CLIP (DP-CLIP) model for fine-grained ZS-SBIR.
Distinct from previous algorithms based on a generalization perspective, our DP-CLIP model provides category-centric insights, enabling flexible adaptation to various target categories and improving retrieval performance. 
2) In our DP-CLIP model, we introduce a visual prompting module and a textual prompting module. These modules utilize several images from the target category and the textual category label, respectively, to guide the model in learning and leveraging category-centric insights. These designs could effectively prompt our model to adapt to the target categories.
3) Through extensive experiments, the effectiveness of our proposed modules is well evaluated. 
Empowered with these designs, our DP-CLIP outperforms the state-of-the-art fine-grained ZS-SBIR method by $7.3\%$ in Acc.@1.

\section{Related Work}

\subsection{ZS-SBIR and Fine-Grained ZS-SBIR}

\textbf{ZS-SBIR.} Sketch-based image retrieval (SBIR) aims to retrieve realistic images with the same category as a given hand-drawn sketch. 
Due to the scarcity and limited variety of sketches available for training, SBIR models inevitably encounter unseen object categories in real-world applications. 
Thus, zero-shot sketch-based image retrieval (ZS-SBIR)~\cite{shen2018zero,yelamarthi2018zero} has been proposed, which aims to construct a robust retrieval model that could handle unseen categories after being trained on seen categories.
Among existing ZS-SBIR methods, some approaches~\cite{liu2019semantic,tian2022tvt,tian2021relationship,wang2022prototype,wang2021transferable} propose to distill generalizable knowledge from pre-trained teacher models into a student retrieval model, which could enhance the generalization capability of the student model. Some other methods~\cite{shen2018zero,dey2019doodle, dutta2019semantically,xuIJCAIprogressive,wang2023cross} utilize additional semantic knowledge to embed sketches and photos into a well-established semantic space. In this space, the potential semantic associations between different categories, such as dogs and cats, could be leveraged to facilitate the transferring of knowledge from seen categories to unseen ones.

\noindent\textbf{Fine-Grained ZS-SBIR.} Existing category-level ZS-SBIR methods struggle with fine-grained retrieval scenarios, such as searching for lost items, which need to match the exact corresponding instance with a given hand-drawn sketch. To address this, fine-grained ZS-SBIR~\cite{pang2019FG_CC} has been proposed, which aims to retrieve photos containing the exact object drawn in the provided sketch. 
This task presents a significant challenge, necessitating that retrieval models could both distinguish instances within the same category and handle varying unseen target categories.
This requires retrieval models to adapt effectively to the target categories and to identify unique discriminative clues within each category for retrieval.
However, existing methods always overlook the need to guide the model in flexibly adapting to target categories but adopt the generalization strategy to acquire knowledge from training set~\cite{pang2019FG_CC,lin2023explainable} or pre-trained models~\cite{sain2023clip} and directly apply this knowledge to unseen categories, which we believe is sub-optimal.
To address this, in this work, we propose a category-specific dual-modal prompting strategy to tailor the model for each target category, thereby enabling it to effectively adapt to the target category and capture unique discriminative clues for improved retrieval.

\subsection{CLIP in Vision Tasks. }
The Contrastive Language-Image Pre-training (CLIP)~\cite{radford2021clip} model has shown remarkable success in a series of zero-shot visual recognition tasks. 
Many recent studies show that the CLIP has a strong generalization capability, facilitating its application in many downstream vision tasks, including fine-grained art classification~\cite{conde2021clip}, image generation~\cite{ramesh2022hierarchical,crowson2022vqgan}, object re-identification~\cite{li2023clip_reid}, 3D point
cloud understanding~\cite{zhang2022pointclip}, video recognition~\cite{lin2022frozen} and retrieval~\cite{deng2023prompt}.
Inspired by these studies, we employ CLIP as our backbone model and make an effort to effectively utilize the pre-existing knowledge inside it.

\subsection{Prompt Tuning Strategy.}
With large pre-trained models like~\cite{radford2021clip,kirillov2023segment} achieving significant advancements in diverse vision tasks, many studies focus on developing effective training methods~\cite{brown2020language,xing2023dual} to efficiently transfer these large models toward downstream tasks. 
Recently, Jia~\etal~\cite{jia2022vptl} have proposed the visual prompt tuning method, which adapts a pre-trained model to different downstream tasks by injecting it with a set of task-specific prompt vectors.
A significant advantage of visual prompt tuning is that it can preserve the pre-existing knowledge inside the parameters of the pre-trained model unchanged while adapting it with the pluggable prompt vectors. 
This inspires us to employ different category-specific prompts to adapt our model to varying target categories.
However, in this step, we notice that the original visual prompts are obtained by tuning a group of learnable vectors on the target categories (task). This could hardly be realized in the fine-grained ZS-SBIR task when handling unseen categories.
Therefore, in our visual prompting module, we employ a small number of images from the target category to flexibly construct category-specific prompts in a generative manner.

\section{Methodology}

\begin{figure*}[]
	\begin{center}\includegraphics[width=1.0\linewidth]{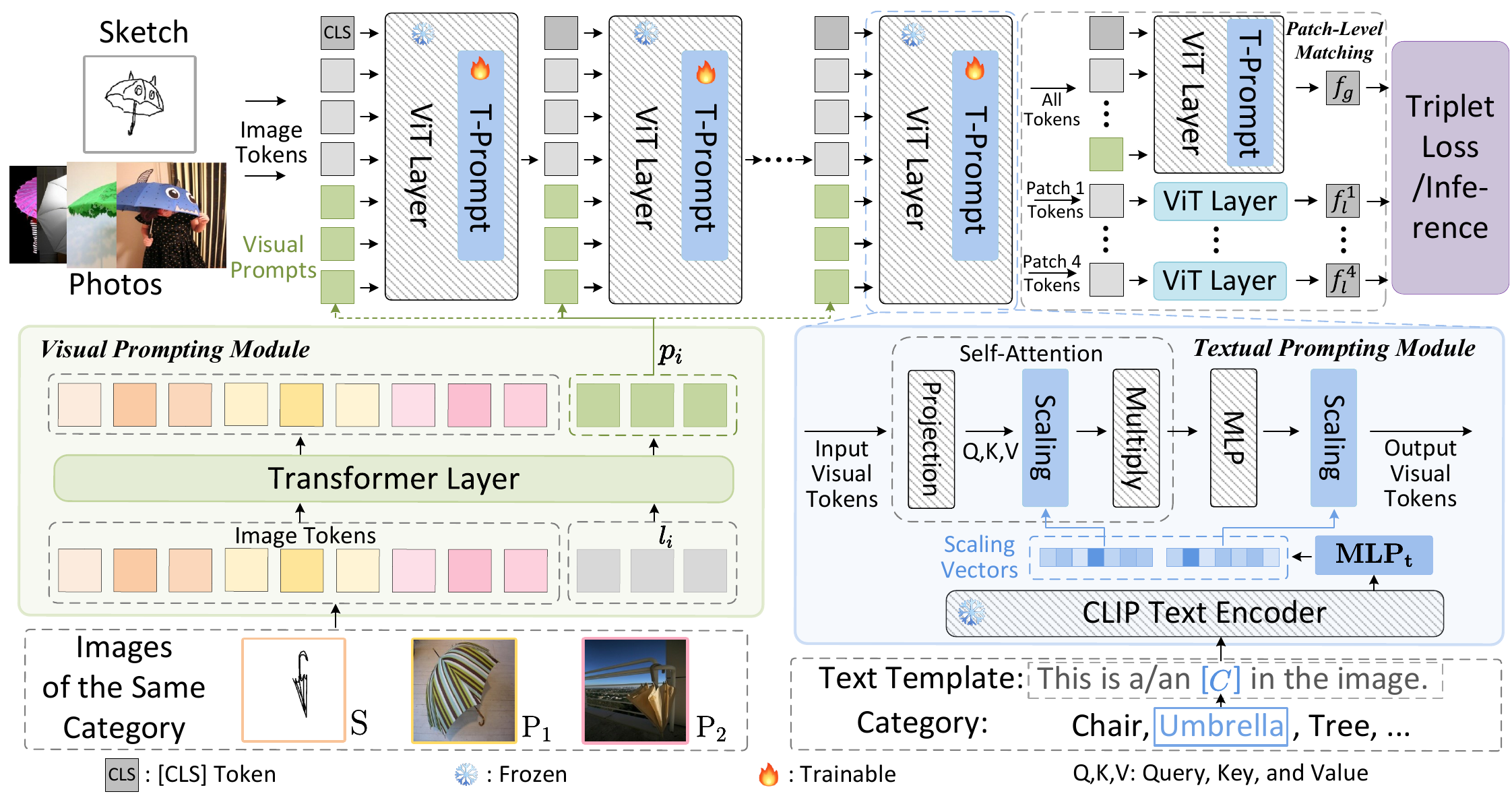}
	\end{center}
	\vspace{-0.7cm}
	\caption{ 
        The architecture of our DP-CLIP model. We use pre-trained CLIP as the backbone model and freeze all parameters, excluding those inside normalization layers. In our DP-CLIP, the visual prompting module is responsible for generating category-specific visual prompts with a set of images from the target category. The textual prompting (T-Prompt) module utilizes the textual category label to produce category-specific channel scaling vectors, guiding our model to adapt to the target category.
		}
    \vspace{-0.4cm}
	\label{fig:framework}
\end{figure*}

\subsection{Problem Definition}

Given a query sketch $\text{S}$, category-level sketch-based image retrieval (SBIR) aims to retrieve photos with the same category from a large photo gallery $\mathcal{G}$ = $\{ \text{P}^i_j \}_{j=1}^{M_i}|_{i=1}^{N_\text{C}}$, which consists ${N_\text{C}}$ categories, and the $i$-th category has $M_i$ photos. 
Zero-shot SBIR (ZS-SBIR)~\cite{shen2018zero} goes one step further by training retrieval models on limited seen categories $\mathcal{C}^s$ = $\{C^s_i\}^{N_s}_{i=1}$ while testing them among unseen categories $\mathcal{C}^u $ = $\{C^u_i\}^{N_u}_{i=1}$, where $N_s$ and $N_u$ denote the number of seen and unseen categories.
Notably, ZS-SBIR is still a category-level retrieval task. 
Recently, a challenging fine-grained ZS-SBIR~\cite{pang2019FG_CC} has been proposed, which aims to retrieve the photo containing exactly the same object as the given sketch from a group of photos within the target category. 
Formally, in each target category $C_i^u$, given a query sketch $\text{S}^i$, fine-grained ZS-SBIR aims to retrieve the corresponding photo with the same identity from the photo gallery $\mathcal{G}^i$ = $\{\text{P}^i_j \}_{j=1}^{M_i}$.  


\subsection{Overall Framework}
In this work, we propose a dual-modal prompting CLIP  (DP-CLIP) model to address the fine-grained ZS-SBIR task. The overall framework of our DP-CLIP is shown in Figure~\ref{fig:framework}.
In our DP-CLIP, the pre-trained CLIP is employed as the backbone model due to its promising pattern recognition capability and abundant knowledge learned from varying pre-training data.
To prevent overfitting and the subsequent loss of accumulated knowledge in CLIP, in our DP-CLIP model, we freeze all parameters of CLIP except the layer normalization (LayerNorm) parameters.
Besides, the core innovation of our model lies in providing category-centric guidance, enabling dynamic adaptation to unseen target categories instead of merely relying on inherent knowledge derived from the training data or pre-trained models.
To achieve this, as shown in Figure~\ref{fig:framework}, we introduce two key components in our DP-CLIP model, namely, the visual prompting module and the textual prompting module.
These two modules are designed to extract category-centric insights from the given support images and the textual label of the target category. 
By integrating these insights into category-specific prompt tokens and channel scaling vectors, our model could adapt effectively to the target category.
Besides, to capture detailed local correspondences between sketches and photos, a customized patch-level matching module is employed on the last vision transformer (ViT) layer of CLIP.

\subsection{Visual Prompting Module}

In this section, we would like to illustrate the visual prompting (V-Prompt) module. One of our major insights inside the V-Prompt module is to strike an optimal balance between retaining the generalizable knowledge embedded in the CLIP model and tailoring it to the ZS-SBIR task and unseen target categories. In this step, we notice the visual prompt tuning (VPT)~\cite{jia2022vptl} could be a good tool that preserves the knowledge contained in the parameters of the CLIP model unchanged and adapts it into downstream tasks by injecting a set of task-specific prompt vectors. 
Although the VPT could help to avoid overfitting and loss of generalizable knowledge, it necessitates providing a group of customized prompt vectors to adapt the model to different target tasks.
In the context of the fine-grained ZS-SBIR task, we could hardly train a set of customized prompt vectors for every target category, especially for the unseen categories.
To address this issue, in our V-Prompt module, we consider constructing the visual prompts in an adaptive generation manner.
Specifically, we propose to use a transformer layer to transfer the category-centric insights from a group of images within the target category into a group of learnable vectors, which are then used as the category-specific visual prompts and injected into the CLIP ViT model. With these category-centric insights, our model could effectively adapt to diverse target categories.

Practically, in our V-Prompt module, we utilize three randomly sampled images (one sketch and two photos) within the target category as support images to generate the category-specific prompts. 
For the training (seen) categories, the support images are randomly selected from the training set.
In terms of test (unseen) categories, the support images are pre-selected from the image gallery randomly, which are also available in real-world retrieval scenarios.
Here, we denote the three support images as $(\text{S}, \text{P}_{1},\text{P}_{2})$.
Firstly, we utilize a convolution layer to extract the features of these images as $(f_{\text{S}},f_{\text{P}_1},f_{\text{P}_2})$.
Thereafter, a transformer layer is employed to transfer the category-centric knowledge from these features into a group of learnable tokens $l\in \mathbb{R}^{N, L_V}$, where $N$ is the number of tokens and $L_V$ indicates the channel dimension.
This step could be written as,
 \vspace{-0.3cm}
\begin{equation} 
 \begin{aligned}
 [\_, \_, \_, p]=\mathrm{Trans}([f_{\text{S}},f_{\text{P}_1},f_{\text{P}_2}, l]),
 \end{aligned}\label{Eq:trans}
  \vspace{-0.1cm}
\end{equation}
where $[]$ indicates the concatenation operation; $\mathrm{Trans}(\cdot)$ denotes the employed transformer layer; $\_$ is the output of support image features, which are discarded; $p$ denotes the category-specific visual prompts that are then injected into the ViT layer of CLIP.
Here, we follow~\cite{jia2022vptl} to adopt deep prompt tuning to adapt our model, which requires an independent set of prompt tokens for each ViT layer.
To construct category-specific prompt vectors for all 12 ViT layers, we use $12$ groups of learnable vectors $\{l_i\}^{12}_{i=1}$ and a shared transformer layer to repeat Equation~\ref{Eq:trans} 12 times. This could generate the prompt tokens $\{p_i\}^{12}_{i=1}$ for each ViT layer. With these prompt tokens, we then integrate them into our DP-CLIP model to adapt to the target category. 
This step could be formulated as, 
 \vspace{-0.3cm}
\begin{equation} 
 \begin{aligned}
 [ \_,h_{i+1}]=\mathrm{ViT}_{i}([p_{i}, h_{i}]),
 \end{aligned}
\end{equation}
where $\mathrm{ViT}_{i}(\cdot)$ is the $i$-th ViT layer inside the CLIP; $\_$ denotes the output of $p_{i}$, which is discarded; $h_{i}$ is the hidden features of the $i$-th ViT layer.

\begin{figure}[]
	\begin{center}\includegraphics[width=1.0\linewidth]{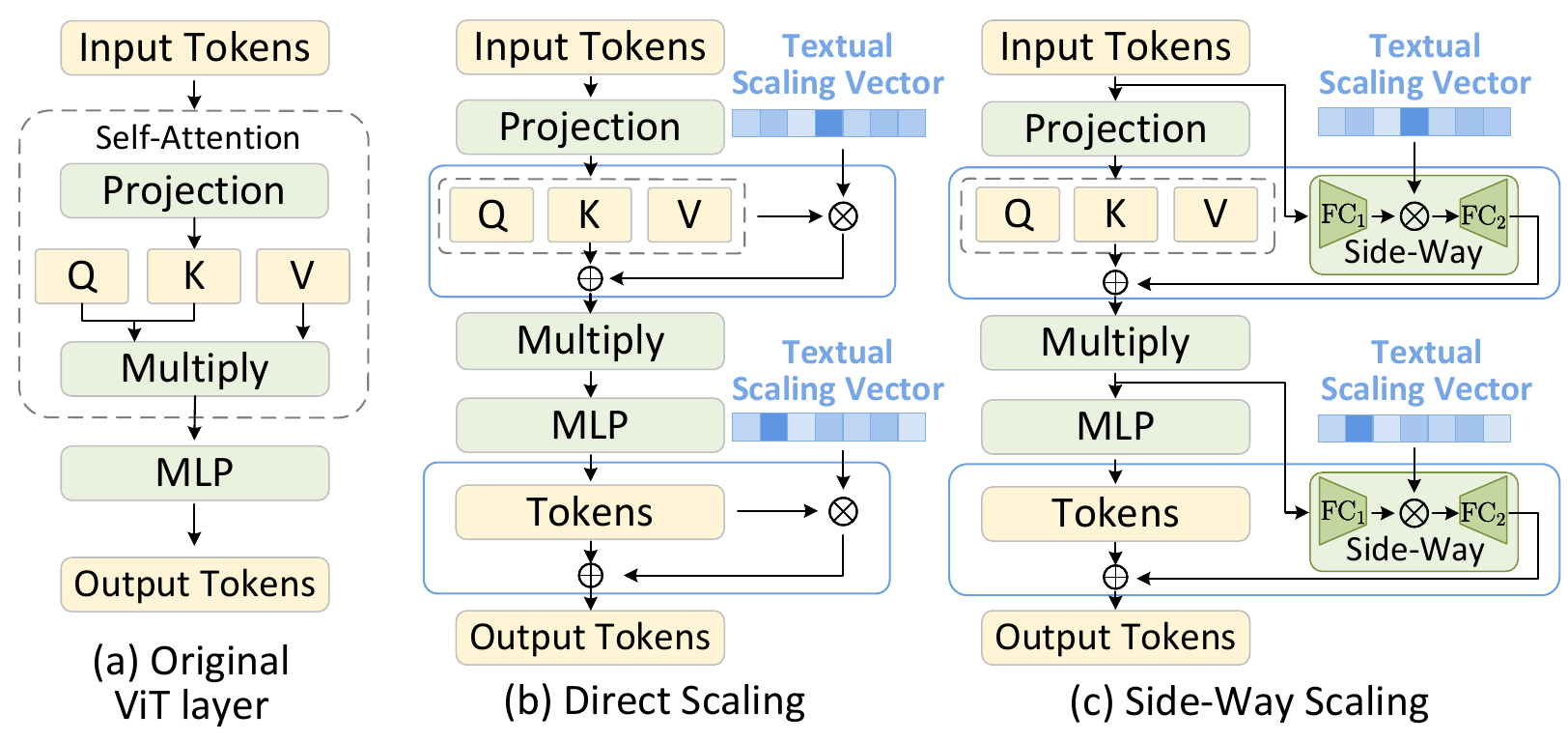}
	\end{center}
	\vspace{-0.7cm}
	\caption{ 
        The architecture of (a) Original ViT layer, (b) Direct Scaling, and (c) Side-Way Scaling.  
		}
    \vspace{-0.5cm}
	\label{fig:text_scale}
\end{figure}

\subsection{Textual Prompting Module} \label{sec:text_prompt}

In this part, we would like to illustrate the textual prompting module. Here, our key insight is to not only exploit the generalization potential of CLIP's visual encoder but also make good use of the generalizable knowledge inherent in its textual encoder. To achieve this, our approach involves using the textual encoder to embed the target category's label to textual features, which are then employed to adapt the visual encoder to the specific target category. 
Given the target category label $C$, the process of extracting textual features $T\in \mathbb{R}^{L_T}$  could be formulated as,
\vspace{-0.2cm}
\begin{equation} 
 \begin{aligned}
 T=\mathrm{Text\_Encoder}(\mathrm{Temp}[C]),
 \end{aligned}
  \vspace{-0.2cm}
 \label{equ:CLIP_text}
\end{equation}
where $\mathrm{Text\_Encoder}(\cdot)$ is the textural encoder of CLIP; $\mathrm{Temp}$ is a textual template formatted as ``This is a/an $[C]$ in the image.'', where $[C]$ is the category label, as shown in Figure~\ref{fig:framework}. To prevent modifications to the structure of the visual encoder that might disrupt its inherent knowledge, in this study, we use the textual features $T$ to produce sets of channel scaling vectors, guiding the adaptation of hidden features inside the visual encoder. As shown in Figure~\ref{fig:framework}, a multi-layer perception ($\mathrm{MLP_{t}}$) module is utilized to produce the category-specific scaling vector.
Regarding the locations for applying the guidance, we implement text-guided scaling over the features produced by both the Projection layer and MLP in each transformer layer inside the visual encoder. 
In this part, we give two approaches for applying the scaling vectors, namely direct scaling and side-way scaling, which we will discuss in detail sequentially. 

\textbf{Direct Scaling}. The direct scaling operation, as shown in Figure~\ref{fig:text_scale} (b), directly multiplies the generated textual scaling vectors over the visual hidden features.
This step could be written as,
\vspace{-0.2cm}
\begin{equation} 
 \begin{aligned}
 V^{\prime}=\mathrm{MLP_{t}}(T)\times V + V,
 \end{aligned}
  \vspace{-0.2cm}
\end{equation}
where $V$ is hidden features; $\times$ denotes the multiplication operation in channel dimension.
Through this straightforward procedure, we could leverage textual category features to guide visual features, encouraging them to focus on channels pertinent to the target category. A major advantage of this approach is that it only increases a simple $\mathrm{MLP_{t}}$ module without disturbing the prior knowledge inherent in the parameters of the visual encoder. Nonetheless, this approach is constrained by its limited flexibility. Specifically, the text guidance could only select and scale pre-existing knowledge within the visual encoder but lacks the capability for flexible calibration to better adapt to the ZS-SBIR task and unseen categories. Therefore, we further introduce the side-way scaling method.

\textbf{Side-Way Scaling.} In the side-way scaling, we 
introduce a side-way module to adapt the visual encoder. Specifically, as shown in Figure~\ref{fig:text_scale} (c), we integrate side-way modules to the Projection layer and MLP inside each ViT layer. In terms of structure, the side-way module consists of two fully connected layers. 
The textual scaling is applied over the hidden features of the side-way modules. Taking the adaptation of the Projection ($\mathrm{Proj.}$) layer as an example, this process could be written as,
 \vspace{-0.2cm}
\begin{equation} 
 \begin{aligned}
 V^{\prime}=\mathrm{FC_{2}}(\mathrm{MLP_{t}}(T)\times \mathrm{FC_{1}}(V)) +  \mathrm{Proj.}(V),
 \end{aligned}
 \vspace{-0.2cm}
\end{equation}
where $\mathrm{FC_{1}}$ and $\mathrm{FC_{2}}$ are fully connected layers inside the side-way module.
Here, $\mathrm{FC_{1}}$ embeds visual features from dimension $L_{V}$ into $L_{S}$, while $\mathrm{FC_{2}}$ embeds the features from dimension $L_{S}$ back to $L_{V}$. 
Through this approach, we could preserve the original knowledge of the Projection layer while leveraging the side-way module to learn a more effective calibration paradigm for the fine-grained ZS-SBIR task and target categories.
This enables textual features to more effectively guide the visual model in adapting to novel categories.
Although this method enhances the effectiveness of text guidance, a potential concern is that the additional side-way module might introduce a significant number of additional parameters, potentially leading to a decrease in model efficiency. 
To address this issue, we consider reducing the hidden feature dimension $L_S$ of the side-way module to compress the model parameters. In this step, we discover that
our text-guided module could also perform well when using extremely low-dimension ($16$) hidden features, only introducing a few additional parameters.

\subsection{Patch-Level Matching Module.} 
In addition to the visual and textual prompting modules, we design a customized patch-level matching module in our DP-CLIP. In this module, we aim to extract features of corresponding local patches inside sketches and photos for comparison, thereby realizing accurate retrieval.
To do so, we divide each sketch or photo into four patches for comparison.
Specifically, we divide features of the penultimate ViT layer, which has the spatial scale of $7\times7$, into four $5\times5$ patches, oriented in four directions: top-left, top-right, bottom-left, and bottom-right (as shown in Figure 1 in the Appendix). The features within each patch are extracted to serve as local features. These local features are then concatenated with the copied [CLS] token features and processed through distinct local ViT layers. The parameters of each local ViT layer are copied from the final ViT layer of CLIP, and only the LayerNorm parameters are optimized. 
The output features of the [CLS] token in each local branch are treated as the final local features for this patch. 

\subsection{Training and Inference} 
In the training stage, for each batch, we randomly select a seen category and sample a set of sketch-photo pairs within this category as input for the DP-CLIP model. 
Particularly, a sketch and two photos are randomly selected to generate the category-specific visual prompts.
Besides, the textual scaling vector is produced according to the textual category label. 
In the last ViT layer, the output features of the [CLS] token are extracted as global features for loss computation. 
Besides, the local features extracted by the patch-level matching module are also employed for loss computation. 
The overall loss $\mathcal{L}$ could be written as following, 
\begin{equation}
\begin{split}
\mathcal{L} = \mathcal{L}_{\text{Tri}}(f_{g}) + 0.1 \sum\nolimits_{k=1}^{4}  \mathcal{L}_{\text{Tri}}(f_{l}^{k}) 
\end{split}
\end{equation}
where $\mathcal{L}_{\text{Tri}}$ is the triplet loss, $0.1$ is the trade-off parameter, and $f_{g}$ is the global features, $f_{l}^{k}$ indicates the local features extracted in the $k$-th local branch.


In the inference stage, we first use the global features to compute the global distance $d_{g}$ between sketches and images. Then we employ the local features to compute four local distances $d_{l}^{k} (k\in \{1,2,3,4\})$. Finally, we add the $d_{g}$ and all $d_{l}^{k}$ together for inference.

\section{Experiment}


\subsection{Experiment Settings.}
\textbf{Datasets.} 1) \textbf{Sketchy}~\cite{sangkloy2016sketchy_dataset} is used for the fine-grained ZS-SBIR experiments, which is the only available multi-category instance-level SBIR dataset. 
This dataset is composed of $125$ categories. Each category contains $100$ photos, and each photo is associated with $5$ to $8$ sketches, resulting in a total of $75,471$ sketches. During the data collection process, each sketch is created based on one specific photo, which provides instance-level correspondences between sketches and photos and makes the dataset suited for the fine-grained ZS-SBIR task. 
In our experiments, we follow~\cite{yelamarthi2018zero} to train on $101$ seen categories and test on $24$ unseen categories.




\noindent2) \textbf{Sketchy Ext}~\cite{liu2017deep}  \textbf{and TUBerlin Ext}~\cite{zhang2016sketchnet} are two popular category-level ZS-SBIR benchmarks. 
The Sketchy Ext dataset extends extra $60,502$ photos to the basic Sketchy~\cite{sangkloy2016sketchy_dataset} dataset, resulting in a total of $73,002$ photos. 
Following~\cite{sain2023clip}, it is split into $104$ training categories and $24$ test categories.
The TUBerlin Ext dataset is collected by drawing sketches and gathering photos based on a given category label. It contains $250$ categories, and each category consists of $80$ sketches and approximately $820$ photos on average. 
In the zero-shot set-up, we follow~\cite{liu2019semantic} to use $220$ categories for training and $30$ categories for inference. 

\noindent\textbf{Evaluation Protocol.} 
For \textbf{fine-grained ZS-SBIR}, we follow the configuration in~\cite{pang2019FG_CC,sain2023clip} to independently apply instance-level retrieval in each test category, and then we compute the average results of all test categories. 
Following~\cite{pang2019FG_CC}, Acc.@1, Acc.@5, and Acc.@10 are reported, which reflect the percentage of sketches having correctly matched photos in the top-1, top-5, and top-10 retrieval lists, respectively. 
For \textbf{category-level ZS-SBIR}, following the standard evaluation protocol, mean average precision in the whole (mAP@all) and top-200 (mAP@200) retrieval results and precision for the top-100 (Prec@100) and top-200 (Prec@200) retrieval results are reported.

\noindent\textbf{Implement Details. }
In our DP-CLIP, we employ the CLIP~\cite{radford2021clip} equipped with ViT-B/32 as the backbone model.
The number of visual prompts inserted into each ViT layer is set to $3$.  
Both sketches and photos are resized to $224 \times 224$. 
During training, photos are augmented with grayscale with a probability of $0.5$, and sketches and photos are simultaneously horizontally flipped with a $0.5$ probability. 
The batch size is set to $64$, and the model is trained for $60$ epochs with the Adam optimizer. 
The learning rate is set to $1e-6$ for the LayerNorm parameters inside CLIP and $1e-5$ for other trainable parameters. 
The margin of the triplet loss is set to $0.15$. 
In this section, the exhibited performance of our DP-CLIP model belongs to the version employing side-way scaling with the hidden dimension of $16$.


\begin{table*}[!htbp]
    \caption{Performance comparison of our method against other methods on the fine-grained ZS-SBIR and category-level ZS-SBIR settings.}
    \vspace{-0.3cm}
    \centering
    \scriptsize
    \setlength{\aboverulesep}{0pt}
    \setlength{\belowrulesep}{2pt}
    \resizebox{0.95\textwidth}{!}{ 
    \begin{tabular}{rl|ccc|cccc}
    \toprule
    & & \multicolumn{3}{c|}{\textbf{Fine-Grained ZS-SBIR}} & \multicolumn{4}{c}{\textbf{Category-Level ZS-SBIR}}\\ 
    \cline{3-9} 
     \multicolumn{2}{c|}{\textbf{Method}} & \multicolumn{3}{c|}{{Sketchy}~\cite{sangkloy2016sketchy_dataset}} & \multicolumn{2}{c}{{Sketchy Ext}~\cite{liu2017deep}} & \multicolumn{2}{c}{{TUberlin Ext}~\cite{zhang2016sketchnet}} \\
    \cline{3-9}  
    & & Acc.@1 & Acc.@5 & Acc.@10 & mAP@200 & Prec@200 & mAP@all & Prec@100 \\ 
    \hline
    CVPR'18 & ZSIH~\cite{shen2018zero} & -- & --  & -- & -- & -- & $0.220$ & $0.291$ \\
    CVPR'19 & SEM-PCYC~\cite{dutta2019semantically} & -- & --  & --  & -- & -- & $0.297$ & $0.426$ \\
    CVPR'19 & CC-DG~\cite{pang2019FG_CC} & $22.60$ & $49.00 $ & $ 63.30$ & -- & -- & -- & --\\ 
    ICCV'19 & SAKE~\cite{liu2019semantic} & -- & --  & --& $0.497$ & $0.598 $ & $0.475$ & $ 0.599$ \\
    AAAI'20 & SketchGCN~\cite{zhang2020zero} & -- & --  & -- & $0.568$ & $0.487$ & $0.324$ & $0.505$ \\
    TPAMI'21 & ZS-TCN~\cite{wang2021transferable} & -- & --  & -- & $0.516$ & $0.608$ & $0.495$ & $0.616$ \\
    AAAI'22 & TVT~\cite{tian2022tvt} & -- & --  & -- & $0.531$ & $0.618$ & $0.484$ & $0.662$ \\
    CVPR'22 & Sketch3T~\cite{sain2022sketch3t} & -- & -- & -- & -- &  -- & $0.507$ & -- \\
    CVPR'23 & ZSE-SBIR~\cite{lin2023explainable} & $17.94$ & $39.93$ & $53.03$ & $0.525$ & $0.624$ & $0.542$ & $0.657$  \\
    CVPR'23 & B-CLIP~\cite{sain2023clip} & $28.68$ & $62.34$ & --  & $0.723$ & $0.725$ & $0.651$ & $0.732$ \\
    \hline
    & DP-CLIP (Ours) & $\mathbf{35.98}$ & $\mathbf{64.50}$ & $\mathbf{76.14}$
    & $\mathbf{0.771}$ & $\mathbf{0.739}$ & $\mathbf{0.663}$ & $\mathbf{0.734}$ \\
    \bottomrule
    \end{tabular}}
    \label{tab:sota}
    \vspace{-0.6cm}
\end{table*}

\subsection{Comparison with State-of-the-Arts. }
\textbf{Fine-Grained ZS-SBIR.}
The performance of our DP-CLIP and state-of-the-art methods on the Sketchy dataset under the fine-grained ZS-SBIR setting is exhibited in Table~\ref{tab:sota}. 
From these results, we can find that DP-CLIP significantly outperforms other algorithms, achieving an improvement of $7.30\%$ in Acc.@1 over the best competitor B-CLIP~\cite{sain2023clip}.
Among the competitors, the CC-DG~\cite{pang2019FG_CC} and ZSE-SBIR~\cite{lin2023explainable} are generalization-based approaches, which aim to learn the sketch-photo correspondences from seen categories and then directly apply the accumulated knowledge to unseen categories.
However, this direct transfer method is sub-optimal as partial knowledge pertaining to seen categories might be ineffective or potentially harmful when applied to unseen categories.
The best competitor, B-CLIP~\cite{sain2023clip}, leverages the robust generalization capabilities of CLIP by utilizing it as the backbone model and adapts it to the SBIR task by leveraging a group of visual prompts shared for all images. 
However, it lacks the necessary flexibility to adapt the pre-existing knowledge inside CLIP to each specific target category.
In contrast, with our designed dual-modal prompting strategy, 
we could effectively inject category-centric insights into the CLIP model, thereby encouraging it to adapt dynamically to the target categories. Empowered by our designs,  our DP-CLIP outperforms the B-CLIP by $7.30\%$ Acc.@1.



\noindent\textbf{Category-Level ZS-SBIR.}
Given that only one fine-grained multi-category SBIR dataset is available, to comprehensively explore the effectiveness of our DP-CLIP model, we further evaluate it under the category-level ZS-SBIR setting.
Under this setting, the retrieval process is conducted among all test categories, aiming to identify photos with the same category as the query sketch.
The overall goal of this setting is to identify the category, which means the category information utilized in our DP-CLIP could not be pre-available. 
Therefore, under this setting, we give two revisions to our dual-modal prompting strategy.
Firstly, we convert the category-specific visual prompts into instance-specific visual prompts, which means we employ each input image as the support image to produce visual prompts for itself.
Secondly, as the category label is not available, we replace the textual template in Equation~\ref{equ:CLIP_text} into ``Focus on the discriminative $[T_p]$", where $[T_p]$ is a learnable text prompt whose output is utilized to generate the scaling vectors. This learnable text prompt is shared for all images, aiming to learn the common discriminative clues benefit for category recognition.
The performance of our DP-CLIP and other ZS-SBIR methods is exhibited in Table~\ref{tab:sota}. 
This demonstrates that, through the aforementioned two simple modifications, our model could efficiently adapt to the category-level retrieval task, yielding promising performance. This indicates the superior recognition and generalization capability of our model.

\subsection{Ablation Study.}
To clarify, in the rest of this section, ``Full Tuning'' indicates directly fine-tuning all the parameters of CLIP; ``Norm Tuning'' denotes only optimizing the LayerNorm parameters inside the CLIP model. All ablation experiments are conducted on the Sketchy dataset.

\begin{table}
  \caption{Ablation studies about the effectiveness of our designed modules.
  V-Prompt, T-Prompt, and Local denote the visual prompting module, textual prompting module,  and patch-level matching module, respectively. Dual-Prompt indicates the version employing both the visual and textual prompting modules.}
  \vspace{-0.2cm}
  \centering
  \resizebox{0.47\textwidth}{!}{ 
  \begin{tabular}{lccc}
    \toprule
    Method  & Acc.@1 & Acc.@5 & Acc.@10 \\ 
     \hline \hline
    Full Tuning &  $25.77$ & $51.84$ & $64.42$ \\ 
    Norm Tuning & $23.96$ &  $50.33$ & $63.58$ \\
    \hline 
    V-Prompt &  $30.29$	& $57.85$ &	$70.80$ \\
    T-Prompt &  $31.11$ & $59.68$ & $72.25$ \\
    Dual-Prompt & $34.33$ & $62.94$ & $75.04$ \\
    \hline
    Dual-Prompt+Local &  $35.98$ & $64.50$ & $76.14$ \\
    \bottomrule
  \end{tabular}}
  \vspace{-0.4cm}
  \label{tab:ablation}
\end{table}

\noindent\textbf{Effectiveness of the Proposed Modules.}
To evaluate the effectiveness of the proposed modules in our DP-CLIP, a series of ablation experiments are conducted on the Sketchy dataset, and the results are listed in Table~\ref{tab:ablation}. 
First, for comparison, we give the performance of two baseline models, \ie, Full Tuning and Norm Tuning.
On the basis of the Norm Tuning, the visual prompting (V-Prompt) module and the textual prompting (T-Prompt) module are employed, respectively. 
From Table~\ref{tab:ablation}, we can find that they bring an improvement of $6.33\%$ and $7.15\%$ in Acc.@1, demonstrating the effectiveness of category-level insights provided by these modules.  
Furthermore, by combining the category-level visual and textual guidance together, an average improvement of $3.63\%$ in Acc.@1 is brought. 
In addition, to capture the local correspondences for fine-grained retrieval, a patch-level matching module (Local) is further utilized, which brings an additional improvement of $1.65\%$ Acc.@1.

\begin{table}
  \caption{Performance comparison between our Category-Specific Visual Prompting (VP) and other VP methods. Common VP denotes employing learnable vectors as prompts shared for all input images, and Instance-Specific VP indicates employing each input image to generate adaptive visual prompts for itself.
  }
  \vspace{-0.2cm}
  \centering
  \resizebox{0.48\textwidth}{!}{ 
  \begin{tabular}{lccc}
    \toprule
    Method  & Acc.@1 & Acc.@5 & Acc.@10 \\ 
     \hline \hline
    Full Tuning &  $25.77$ & $51.84$ & $64.42$ \\ 
    \hline
    Common VP &  $27.30 $  & $ 54.78 $  & $ 68.25$ \\ 
    Instance-Specific VP &  $29.56$  & $57.29$  & $70.30$ \\
    Category-Specific VP &  $30.29$	& $57.85$ &	$70.80$ \\ 
    \bottomrule
  \end{tabular}}
  \vspace{-0.5cm}
  \label{tab:ablation_v}
\end{table}

\noindent\textbf{Effectiveness of Category-Specific Visual Prompts.} 
In the V-Prompt module of our DP-CLIP, we utilize a set of images within each target category to generate category-specific visual prompts, facilitating efficient adaptation to the target category.
Here, a concern may arise regarding whether the effectiveness of our V-Prompt module stems from the category-centric insights it incorporates or is merely a result of employing the prompt tuning technology~\cite{jia2022vptl}.
To respond to this concern, we compare our category-specific visual prompting (Category-Specific VP) with other variants, including 1) Common VP that employs learnable vectors as prompts shared for all input images and 2) Instance-Specific VP that employs each input image to generate adaptive visual prompts for itself.
These two compared variations both adopt deep prompt tuning as our Category-Specific VP.
As shown in Table~\ref{tab:ablation_v}, Common VP only brings $1.53\%$ Acc.@1 improvement to the Full-Tuning model, which has a $2.99\%$ Acc.@1 gap from our employed Category-Specific VP. 
This result is expected since the common prompts could only accumulate shared knowledge over the seen categories, failing to guide the model to adaptively capture the unique discriminative clues within each specific target category.   
Additionally, Instance-Specific VP brings $3.79\%$ Acc.@1 improvement to the Full-Tuning model. 
Although Instance-Specific VP yields the necessary flexibility for adaptation, it only employs one single image to learn target category insights, which could be insufficient, thereby leading to a $0.73\%$ Acc.@1 gap to our Category-Specific VP. 
In comparison, our V-Prompt module employs three support images within the target category to extract category-centric insights and employ them to guide our model to dynamically adapt to each target category, thereby surpassing all compared variants. 
These results indicate that the effectiveness of our V-Prompt module does not merely stem from employing prompt tuning technology but the valuable category-centric insights it brings.

\begin{table}
\caption{Ablation studies of different scaling strategies in our T-Prompt module. Side-Way indicates the model only employs the side-way module but discards the textual scaling. 
  }
  \vspace{-0.2cm}
  \centering
  \begin{tabular}{lccc}
    \toprule
    Method  & Acc.@1 & Acc.@5 & Acc.@10 \\ 
     \hline  \hline
    w/o T-Prompt &  $30.29$	& $57.85$ &	$70.80$ \\
    \hline
    Direct Scaling & $33.07$ &	$61.53$ & $73.47$ \\ 
    Side-Way & $31.98$ & $59.84$ & $72.99$  \\ 
    Side-Way Scaling &  $34.33$ & $62.94$ & $75.04$ \\ 
    \bottomrule
  \end{tabular}
  \vspace{-0.7cm}
  \label{tab:ablation_text}
\end{table}

\noindent\textbf{Ablation about the Textual Scaling.} 
In our DP-CLIP, we
employ the textual category label to generate a set of channel scaling vectors and then apply them to the visual hidden features. 
As we mentioned in Section~\ref{sec:text_prompt}, two kinds of scaling strategies, \ie, direct scaling and side-way scaling, are introduced (for more details please refer to Figure~\ref{fig:text_scale}).
Here, we give some ablation experiments to explore the effectiveness of these two strategies, respectively.
The experimental results are given in Table~\ref{tab:ablation_text}, from which we can find that the direct scaling could bring $2.78\%$ Acc.@1 improvement to our model, which indicates our idea of deriving knowledge from the textual encoder of CLIP to guide the adaptation of visual encoder is feasible and effective.  
Besides, if adopting the side-way scaling strategy, the performance could be further improved by $1.26\%$ Acc.@1.
The reason behind this improvement could be that the side-way module provides more flexibility, 
which allows the T-Prompt module to learn a more effective calibration paradigm to adapt our model to the unseen categories.
In addition, for comparison, we also give the performance of the version only employing the side-way module but discarding the textual scaling (``Side-Way''), which suffers a significant $2.35\%$ Acc.@1 gap from our side-way scaling strategy. 
This result indicates that the effectiveness of our T-Prompt module is caused by not merely introducing additional parameters but the textual category guidance.

\begin{figure}[]
	\begin{center}\includegraphics[width=1.0\linewidth]{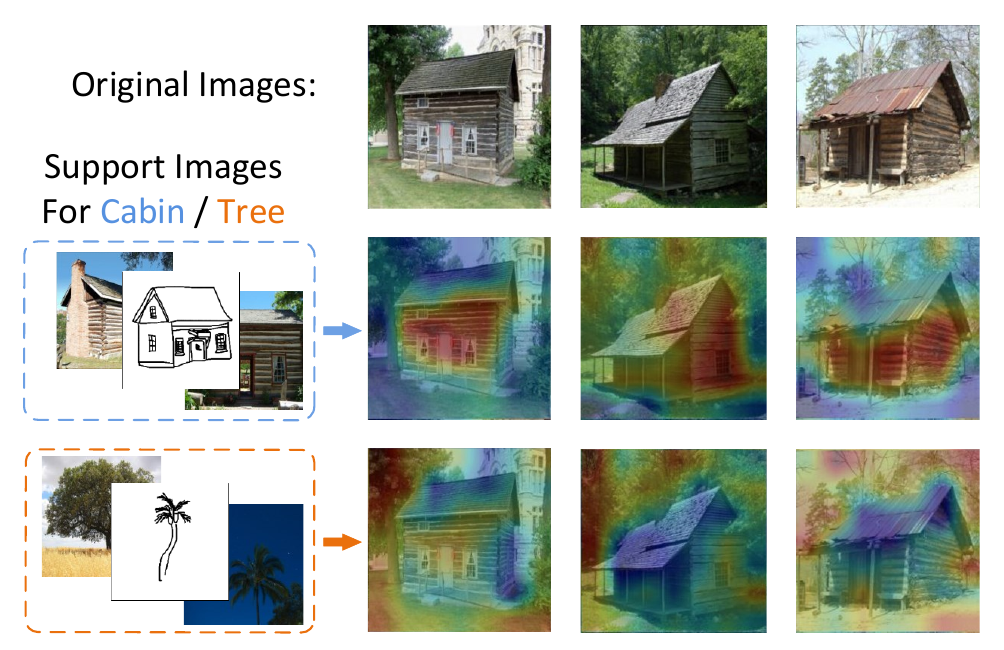}
	\end{center}
	\vspace{-0.6cm}
	\caption{ 
        Visualization results of our visual prompting module. We analyze and display the information within the category-specific visual prompts by computing the cosine similarity between the visual prompts and image tokens. Regions of high similarity are highlighted in red, while regions of low similarity are indicated in blue. In this part, we use support images from the ``cabin'' and ``tree'' categories for guidance.
        The results show that
        the visual prompts generated with support images from different categories contain different category insights.
        Given the same images, it could potentially encourage our model to focus on target objects of different categories, thereby efficiently adapting to the target category and improving retrieval.
		}
    \vspace{-0.6cm}
	\label{fig:visualize}
\end{figure}

\noindent\textbf{Visualization Cases.} 
To intuitively illustrate how our category-specific visual prompting strategy guides the model to adapt to target categories, in Figure~\ref{fig:visualize}, we give some visualization cases of our visual prompting module. As shown in this figure, the visual prompts generated with support images from different categories, \ie, cabin and tree, contain different category insights. Given the same images, it could potentially encourage our model to focus on target objects of different categories, thereby efficiently adapting to the target category and improving retrieval.

\section{Conclusion}
In this work, we propose a DP-CLIP model for fine-grained ZS-SBIR.  Unlike previous methods based on the generalization perspective, our DP-CLIP model provides category-centric insights, allowing flexible adaptation to different target categories for improved retrieval performance. 
In the DP-CLIP model, we introduce a visual prompting and a textual prompting module, which leverage several images from the target category and the text category label to guide the model to learn and apply category-centric insights. 

%

{
    \small
    \bibliographystyle{ieeenat_fullname}
    \bibliography{main}
}

\appendix

\begin{figure}[]
	\begin{center}\includegraphics[width=1.0\linewidth]{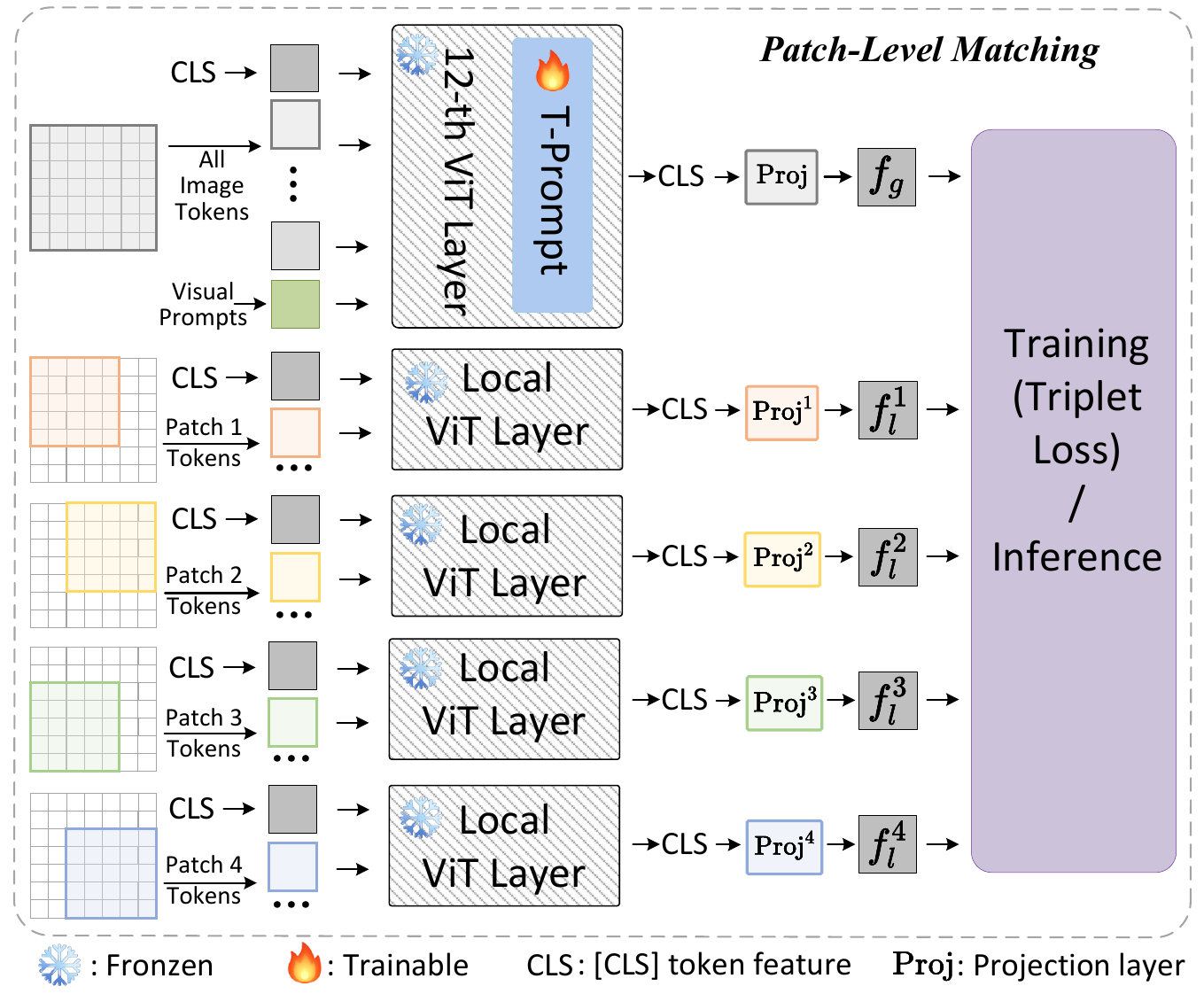}
	\end{center}
	\caption{ 
        The architecture of our patch-level matching module. 
        As shown in the left part, we divide features extracted from the penultimate ViT layer inside CLIP, which has the spatial scale of $7\times7$, into four $5\times5$ patches, oriented in four directions: top-left, top-right, bottom-left, and bottom-right. The feature tokens within each patch are concatenated with a copied [CLS] token feature and then processed through a local ViT layer. The output features of the [CLS] token in each local branch are sent into a linear projection layer, whose outputs are treated as the final local features for this patch. 
		}
	\label{fig:patch}
\end{figure}

\section{Details of Patch-level Matching Module}
In this section, we would like to introduce the details of our employed patch-level matching module. In the relevant fine-grained recognition tasks like re-identification, many works~\cite{sun2018beyond, he2021transreid} are devoted to capturing and extracting local features from crucial object regions for detailed comparison. Inspired by these works, we designed the patch-level matching module to extract features of corresponding local patches inside sketches and photos for comparison, thereby realizing accurate retrieval. 
To do so, we divide each sketch or photo into four patches for comparison.
Specifically, we divide features extracted from the penultimate ViT layer inside CLIP~\cite{radford2021clip}, which has the spatial scale of $7\times7$, into four $5\times5$ patches, oriented in four directions: top-left, top-right, bottom-left, and bottom-right, as shown in Figure~\ref{fig:patch}. The feature tokens within each patch are extracted as local features. Thereafter, each group of local features is concatenated with a copied [CLS] token feature and then processed through a local ViT layer. The output features of the [CLS] token in each ViT layer are sent into a linear projection layer, whose outputs are treated as the final local features for this patch.

\begin{table*}
  \centering
  \begin{tabular}{l|cccc|ccc|cc}
  \toprule
    Method & V-Prompt & T-Prompt & Dim. & Local & Acc.@1 & Acc.@5 & Acc.@10 & FLOPs(G) & Params(M) \\ 
     \hline 
    Full Tuning & $\times$ & $\times$  & -- & $\times$ & $25.77$ & $51.84$ & $64.42$ & $4.98$ & $100.80$ \\
    V-Prompt & $\checkmark$ & $\times$ & -- & $\times$ & $30.29$	& $57.85$ &	$70.80$ & $5.24$ & $100.83$ \\
    Dual-Prompt & $\checkmark$ & $\checkmark$ & $16$ & $\times$ & $34.33$ & $62.94$ & $75.04$ & $5.27$ & $101.42$ \\ 
    DP-CLIP & $\checkmark$ & $\checkmark$ & $16$ & $\checkmark$ &  $35.98$ & $64.50$ & $76.14$ & $6.01$ & $114.41$ \\  
    \hline
    Dual-Prompt (D) & $\checkmark$ & $\checkmark$ & -- & $\times$ &  $33.07$ &	$61.53$ & $73.47$ & $5.99$ & $100.85$ \\ 
    Dual-Prompt & $\checkmark$ & $\checkmark$ & $16$ & $\times$ & $34.33$ & $62.94$ & $75.04$ & $5.27$ & $101.42$ \\ 
    Dual-Prompt & $\checkmark$ & $\checkmark$ & $64$ & $\times$ & $35.15$ & $63.29$ &  $75.21$ & $5.37$ & $103.19$  \\ 
    Dual-Prompt & $\checkmark$ & $\checkmark$ & $256$  & $\times$ & $35.76$ & $64.10$ & $75.52 $ & $5.82$ & $110.28$ \\
    \bottomrule
  \end{tabular}
  \caption{Analysis about Model Capacity. In this part, we incrementally integrate our designed modules, specifically the visual prompting module (V-Prompt), the textual prompting module (T-Prompt), and the patch-level matching module (Local) into the CLIP ViT model to explore the computational consumption and parameters they bring.
  Besides, in the second block, we conduct an ablation study to assess how the hidden feature dimension of the side-way module in T-Prompt influences our model’s performance and efficiency. The ``Dim.'' indicates the hidden feature dimension of the side-way module inside the T-Prompt. ``Dual-Prompt (D)'' indicates the version employing direct scaling in the T-prompt.
  }
  \label{tab:cap}
\end{table*}


\section{Analysis about Model Capacity}
In this part, we would like to explore the computational consumption and parameters associated with our proposed modules. To achieve this, we incrementally integrate our designed modules, \ie, the visual prompting module (V-Prompt), the textual prompting module (T-Prompt), and the patch-level matching module (Local), into the CLIP ViT model.
Additionally, we conduct an ablation study to assess how the hidden feature dimension of the side-way module in T-Prompt influences our model's performance and efficiency. 
Notably, for each target category, it is only necessary to predict the category-specific visual prompts and textual scaling vectors once, which will be reused for all images. 
Therefore, the computational and parameter requirements of the prediction module could be neglected.
The results are given in Table~\ref{tab:cap}, from which we can find that our designed V-Prompt, T-Prompt, and Local, respectively, bring $0.26$, $0.03$, and $0.74$ additional GFLOPs to the CLIP ViT (Full Tuning) model.
Given that the increases of GFLOPs are an order of magnitude lower than the total computation of the baseline Full Tuning model ($4.98$), while the improvement brought by these modules are significant, \ie, $10.22\%$ Acc.@1, it is reasonable to render them acceptable. 
Moreover, the ablation study on the hidden feature dimension of the side-way module indicates the potential for further enhancing our model's performance by increasing this dimension.

    

\end{document}


\maketitle

\section{Ablation Experiments}

\begin{table}
  \centering
  \resizebox{0.5\textwidth}{!}{ 
  \begin{tabular}{lccc}
    \toprule
    Method  & Acc.@1 & Acc.@5 & Acc.@10 \\ 
     \hline
    CS Visual Prompt &  $\mathbf{29.60}$ & $\mathbf{57.41}$ & $\mathbf{70.58}$\\
    Sketch Gen. CS Prompt &  $28.84$ & $56.45$ & $69.51$\\
    Photo Gen. CS Prompt &  $29.57$ & $57.18$ & $70.15$ \\
    Random Images Gen. CS Prompt &  \\
    
    Unmatched CS Prompt& \\
    \bottomrule
  \end{tabular}}
  \caption{Ablation studies of different category-specific (CS) visual prompts generation methods on the Sketchy dataset. 
  }
  \label{tab:app_ab_v}
\end{table}

\begin{table}
  \centering
  \begin{tabular}{lccc}
    \toprule
    Method  & Acc.@1 & Acc.@5 & Acc.@10 \\ 
     \hline 
    w/o CTFS &  $30.29$	& $57.85$ &	$70.80$ \\
    \hline
    Rank=16 & $34.03$ & $62.34$ & $74.5$  \\ 
    Rank=64 & $34.54$ & $62.92$ &  $74.93$ \\ 
    Rank=256 & $34.92$ & $63.42$ & $75.46 $ \\
    Rank=768 &  $35.77$ & $63.90$ & $75.60$ \\
    \bottomrule
  \end{tabular}
  \caption{Ablation studies about the different ranks of our category text-guided feature scaling (CTFS) component on the Sketchy dataset. CT and LR represent category text-guided scaling and low-rank feature generation. }
  \label{tab:text_rank}
\end{table}

{
    \small
    \bibliographystyle{ieeenat_fullname}
    \bibliography{main}
}
